# Monolingually-Derived Phrase Scores for Statistical Machine Translation Using Neural Networks Vector Representations

Amir Pouya Aghasadeghi

Human Language Technology Lab, Amirkabir University

Tehran, Iran

aghasadeghi@aut.ac.ir

Mohadeseh Bastan

Human Language Technology Lab, Amirkabir University

Tehran, Iran

m.bastan@aut.ac.ir

*Abstract*—In this paper, we propose two new features for estimating phrase-based machine translation parameters from mainly monolingual data. Our method is based on two recently introduced neural network vector representation models for words and sentences. It is the first time that these models have been used in an end to end phrase-based machine translation system. Scores obtained from our method can recover more than 80% of BLEU loss caused by removing phrase table probabilities. We also show that our features combined with the phrase table probabilities improve the BLEU score by absolute 0.74 points.

## I. INTRODUCTION

Statistical machine translation systems are currently trained from large amounts of parallel corpora. These corpora are used for learning parameter of statistical models [1]. Obtaining these parallel corpora are expensive and time-consuming. On the contrary, monolingual data are available for most of the languages and can be found from many different resources easily. Due to these reasons finding a method to train machine translation model with monolingual data instead of bilingual data has been a focus of many studies in the last couple of years.

The idea of the possibility of learning translation model from monolingual corpora came from the similarities and regularities existing between different languages. Contextual similarities [2], orthographic similarities [3, 4], temporal similarities [5] and topic models [6] are some examples of those features. It has been demonstrated [7] that by using these features alone for estimating translation model parameters most of BLEU loss is recoverable.

Another recent path of work in the field of natural language processing is learning continuous vector representations for words, sentences and documents using neural networks. These vector representations can capture significant amounts of syntactic and semantic features between words and sentences. Also, it has been shown by using a linear transformation matrix learned from a small bilingual dictionary, these models can be used for translating words and can extend the seed dictionary [8].

Machine translation system consists of several different parts (e.g. Language models, reordering model, word alignment, phrase extraction and phrase scoring). In this work, we focused on phrase scoring. We use distributed representation neural network models, trained on monolingual corpora, to re-estimate phrase table scores. We convert all the phrases in the phrase table and all the words in the lexical table, to a vector space. Afterward by multiplying a transformation matrix to source vectors, we calculate cosine similarities between vector pairs in projected source space and the target space.

We use these new scores in an end to end phrase-based statistical machine translation system [9]. Similar to [7] work, we assume the phrase table is available to us, but all of its scores have been removed. We show if we only use new proposed scores obtained from monolingual data, we could recover most of the BLEU loss caused by removing the scores from the phrase table. Also by combining them with original phrase table scores, we can improve the baseline machine translation system. In contrast with orthographic features, since both of these models are language independent our method is useful for related and unrelated language pairs, and since none of these models need any extra metadata such as time stamps, this method can be used with any type of monolingual corpora. The only bilingual data required for our method is a small dictionary and a small amount of parallel sentences for training the transformation matrices needed for projecting the source space to the target space. Our scores can be calculated relatively faster than probabilities in statistical machine translation systems. Also, since we do not need to calculate phrase pairs co-occurrence, our method requires much less bilingual data.

## II. BACKGROUND

In machine translation, neural networks were first used by [10, 11]. They used a neural network for example-based machine translation. By now many attempts have been made to improve machine translation using neural networks.

Word representation by using a continuous vector was first done by [12]. Bengio also used a feedforward neural network to learn word representation whose work was followed by many others [13].

Mikolov [14] proposed a new architecture for learning distributed representation of words. In their work, a word vector was learned by using a feedforward neural network. They offered two neural networks for learning word vectors. In CBOW model, a feedforward neural network with an input layer, a projection layer, and an output layer were proposed. The projection layer is common among all words. Thus, the input of this neural network is a window of $n$ future words and $n$ history words of the current word. All the words are projected to a common space. By averaging these vectors, the current word is predicted. In Skipgram model, the input is a word which is fed into the projection layer and the output is $2*n$ vectors for $n$ future and $n$ history words of the current word. As a result, in this model the attempt is to maximize word classification according to the word's neighborhood in a common sentence.

Simultaneously, they also tried to explore similarities among languages by using a transformation matrix between word vectors [8]. They first created two models for source and target languages by using large monolingual datasets, then with a small bilingual dictionary, the linear projection among languages was learned. For this purpose, dictionary's words with different lengths were projected to fixed length vectors by using CBOW or Skipgram model. In practice, when the monolingual data are limited, the Skipgram shows a better representation of words, but CBOW is faster and is recommended for larger datasets. Word vectors are computed with a projection matrix $W$ by using equation (1), where $x_i$ is the vector of the source word $i$ and $z_i$ is the projection of the vector representing the target space.

$$Wx_i = z_i \qquad (1)$$

Finding projection matrix ($W$) in his equation can be viewed as an optimization problem and it can be solved by minimizing the error rate using a gradient descent approach. Their results show that this method covers 92% of English to Spanish dictionary for words. Another advantage of this method is that in can be used for retrieving dictionary's missing words.

The main weakness of the Skipgram and CBOW models is that they can only be used on words and short phrases. [15] proposed a distributed representation of paragraphs. This model learned paragraph vector with fixed lengths from a variable length text (e.g. phrases, sentences, paragraphs, and documents.). The word in the paragraph is predicted with this vector. This work is done in two steps: first the word vector is learned by using a feedforward neural network in CBOW or Skipgram model, then the word vectors are fed into another neural network with paragraph vector with the same length. All the vectors are concatenated or averaged and the output is the new paragraph vector.

### III. RE-ESTIMATING PHRASE TABLE SCORES USING VECTOR REPRESENTATIONS SIMILARITY

Phrase tables usually have four feature scores for translation source language(e) to target language(f).

- Direct phrase translation probability φ(e|f)
- Direct lexical weighting lex(e|f)
- Inverse phrase translation probability φ(f|e)
- Inverse lexical weighting lex(f|e)

In this section, we describe how each of these scores can be replaced by using our monolingual features. The results of our experiments can be found in section (V).

#### A. Lexical weightings

The main purpose of lexical weightings is to reduce the impact of phrase scores which has been overestimated by Maximum Likelihood Estimation (MLE) [7]. Since lexical weighting is computed using word translation probability, its calculation depends on bilingual data. Vectors obtained using CBOW or Skipgram models capture significant linear regularities between similar or related words. These regularities can be found in vectors trained in different languages from comparable corpora. Because of these regularities, if a correct linear transformation matrix is applied to source vectors, it can project them to the target space. It is shown that the closest target vector to the projected source vector is the most probable translation of that word.

In our model, we convert all of the source words and target words existing in the lexicon table to vectors. Then we need a transformation matrix that can project those vector spaces to each other. This matrix can be found by selecting some pairs of words from the source and target side. We train our matrices using most common words in the source side and their translation. Even though one transformation matrix and its inverse should be enough for our purpose, we decided to train two different matrices, one for direct lexical weightings and one for inverse lexical weightings. In this way, our model is more fault-tolerant.

After projecting source vectors and target vectors to each other, we calculate cosine similarity between each of the pairs existing in lexicon table (Equation 2). In this equation, Z is the projection vector of the source vector, and Y is the target vector.

$$similarity = \frac{Z \cdot Y}{\|Z\| \|Y\|} \qquad (2)$$

Because of properties of these models, we have got to the conclusion that word translation probabilities can be replaced by the cosine similarities between vectors. Our results in sections (V) show that our assumption is correct.

By having word translation scores, computing lexical weightings for each of the phrases is a straightforward task. For this purpose, we use equation (3) [9].

$$P_w(f \mid e, a) = \prod_{i=1}^{n} \frac{1}{|\{j \mid (i,j) \in a\}|} \sum_{\forall (i,j) \in a} w(f_i \mid e_j) \qquad (3)$$

The only difference between our model and bilingual lexical weightings is that since our model cannot compute null alignment probability for different words we use a small constant instead of it.

#### B. Phrase probabilities

CBOW model and Skip-gram model cannot work with text inputs of different lengths. Since phrases have variable lengths, we need to use a different model to convert phrases to vectors. We find PV-DM suitable for this task.

Same as the previous part we convert all of the unique phrases in the phrase table to vectors. The next step would be to find a method to project source space and target space to each other. Since PV-DM shares many properties from CBOW and Skip-gram models, the phrase vector spaces can be mapped to each other using the same method described in part (III).

Selecting the right data for training transformation matrices has been a challenge. There have been three options available for training the transformation matrices; word pairs from a dictionary, high probability phrases from the phrase table or short sentences from a parallel corpus. Since phrase pairs in phrase table are found with heuristics, even selecting high probability phrase pairs could be inaccurate. Word vector space and phrase vector space are different and hence using word pairs for transformation matrix training did not yield promising results. Thus, we decide to use a small amount of short sentences from parallel corpora to train our transformation matrices. Same as the previous part, we train two different transformation matrices for the direct phrase score and inverse phrase score. We also calculate the phrase scores using the cosine similarity between vectors.

## IV. EXPERIMENT SETUP

For our experiments, we use Spanish as the source language and English as the target language. We train our phrase-based statistical machine translation model using Moses system [16] on the full version of the parallel Europarl V5 [17]. We limit the maximum phrase length to 6 and remove the lexical reordering feature from Moses training. Limiting the maximum phrase length will reduce the BLEU score of the machine translation system, but since the number of suggestions drops significantly for longer phrases, the phrase probabilities will lose their importance. Our primary purpose of this work is to re-estimate the phrase scores, so we believe these settings is more appropriate. We use default Moses settings for all other parameters. In this way, we can compare our model trained with monolingual data against the parallel system. Also, we train our 4-gram language model using KenLM [18] on the full English Wikipedia combined with Europarl English side. As our test and development sets we use WMT07, each set contained 2000 sentences.

For training CBOW model, we use English and Spanish Wikipedia. The size of these datasets can be found in the Table (1). Our vector space for word and phrase are 200-dimensional. We have trained each of these models with 30 epochs.

|  | # of Words | # of Unique Words |
|---|---|---|
| **English** | 1,693M | 8M |
| **Spanish** | 472M | 3M |

Table 1: Size of Spanish and English Wikipedia, which we used as our monolingual data

We train four transformation matrices for projecting words and phrases vectors from the source side to target side and vice versa. For training word transformation matrices, we use 10000 most common words in Spanish Wikipedia and their translation. For phrase vector transformation matrices, we use 5000 unique randomly selected short sentences (with lengths of 1 to 8) from Europarl corpus. We use the same method described in section (II) for training our transformation matrices. For tuning model weights, in both Moses based system and vector based system we use minimum error rate training [19].

Even though our model needs parallel data for finding phrase pairs, all other bilingual obtained scores have been removed from the phrase table for our experiences. The phrase-table extracted by Moses has near 20 million phrase pairs. On average, for every unigram Spanish phrase, there is near 100 English phrase suggestions. This value reduced to 27 for bigram phrases and 13 for trigram phrases. Therefore phrase score, play a major role in machine translation systems. More results about the impact of phrase scores can be found in the next section.

## V. EXPERIMENT RESULTS

We replace phrase tables scores with four independent monolingual feature scores obtained from our vector-based models. We run our end-to-end statistical machine translation system for each of those scores and their combinations. Also, we combine the monolingual scores with original phrase-table scores by adding them to the phrase-table and then we show the results in term of the BLEU score for combined systems. Table (2) shows our experiments results. The BLEU score of the baseline system that is trained by Moses default settings is shown in the first row. In the second experiment, the result of removing all original phrase scores in the phrase-table is shown and of course, a significant drop in BLEU score is observed. In the third experiment, we replace the direct phrase translation probability with our model direct vector-based similarity. We explore our inverse phrase probability score impact in the fourth experiment. In the next experiment, we add our direct monolingual lexical weightings score to the phrase table of the third experiment. In the sixth experiment, we replace all bilingual phrase table score with our monolingual alternatives scores. In the seventh experiment, the result of combining our proposed scores with original phrase-table scores is shown

| #Exp. | Direct Scores | | Inverse Scores | | BLEU |
|---|---|---|---|---|---|
|  | **Φ** | **Lex** | **Φ** | **Lex** |  |
| **1** | B | B | B | B | 22.48 |
| **2** | - | - | - | - | 6.03 |
| **3** | M | - | - | - | 16.93 |
| **4** | M | - | M | - | 17.7 |
| **5** | M | M | - | - | 18.27 |
| **6** | M | M | M | M | **18.44** |
| **7** | B+M | B+M | B+M | B+M | **23.22** |

Table 2: Our result summary, here 'M' stands for monolingual scores and 'B' stands for Bilingual scores.

As we have shown our method successfully, recover more than 82% percent of the BLEU loss caused by removing bilingual scores from the phrase table. Also, combining our features improve the baseline system BLEU score by 0.74 points.

## VI. CONCLUSION

In this paper, we introduce two new features to approximate phrase table scores. These features can be obtained without using any additional data or metadata such as timestamps, and also, they are language independent. Since these scores do not need phrase pairs, co-occurrences count they can be calculated relatively faster than phrase-probabilities. We show that by using these scores alone, more than 82% of the BLEU loss caused by removing the phrase table scores are recoverable. Also by combining our features with bilingual obtained phrase-table scores, we improve the BLEU score by absolute 0.74 percent.